\documentclass[letterpaper, 10 pt, journal, twoside]{IEEEtran}
\usepackage{amsmath,amsfonts}
\usepackage{algorithmic}
\usepackage{array}
\usepackage{textcomp}
\usepackage{stfloats}
\usepackage{url}
\usepackage{verbatim}
\usepackage{graphicx}
\def\BibTeX{{\rm B\kern-.05em{\sc i\kern-.025em b}\kern-.08em
				T\kern-.1667em\lower.7ex\hbox{E}\kern-.125emX}}
\usepackage{balance}

\usepackage{graphics} 
\usepackage{epsfig} 
\usepackage{float}
\usepackage{amsmath} 
\usepackage{array}
\usepackage{multirow}
\usepackage{cases}
\usepackage{txfonts}
\usepackage{mathptmx} 
\usepackage{times} 
\usepackage{amssymb}  
\usepackage{longtable}
\usepackage{subfigure}
\usepackage{epsfig} 
\usepackage{bigstrut}
\usepackage{cite}
\usepackage{graphicx}
\usepackage{siunitx}
\usepackage{booktabs}
\usepackage[colorlinks=true,linkcolor=black,citecolor=green,urlcolor=blue]{hyperref}

\usepackage[linesnumbered,boxed,ruled,commentsnumbered]{algorithm2e}
\usepackage[capitalize]{cleveref}
\crefname{section}{Sec.}{Secs.}
\Crefname{section}{Section}{Sections}
\Crefname{table}{Table}{Tables}
\crefname{table}{Tab.}{Tabs.}
\usepackage{amsmath, bm}
\usepackage{amssymb}
\usepackage{bbding}

\usepackage{underscore}
\usepackage{makecell}
\usepackage{steinmetz}
\usepackage{tikz}
\usepackage{makecell}

\definecolor{lime}{HTML}{A6CE39}
\DeclareRobustCommand{\orcidicon}{
	\begin{tikzpicture}
		\draw[lime, fill=lime] (0,0)
		circle[radius=0.16]
		node[white]{{\fontfamily{qag}\selectfont \tiny \.{I}D}}; 
	\end{tikzpicture}
	\hspace{-2mm}
}

\foreach \x in {A, ..., Z}{%
	\expandafter\xdef\csname orcid\x\endcsname{\noexpand\href{https://orcid.org/\csname orcidauthor\x\endcsname}{\noexpand\orcidicon}}
}

\begin{document}
\title{LinK3D: Linear Keypoints Representation for 3D LiDAR Point Cloud}

\author{Yunge Cui\hspace{-1.5mm}\orcidA{}, Yinlong Zhang\hspace{-1.5mm}\orcidB{}, Jiahua Dong\hspace{-1.5mm}\orcidC{}, Haibo Sun\hspace{-1.5mm}\orcidD{}, Xieyuanli Chen\hspace{-1.5mm}\orcidE{}, Feng Zhu\hspace{-1.5mm}\orcidF{} 
	\thanks{Manuscript received: September 18, 2023; Revised: December 1, 2023; Accepted: December 26, 2023. 
		This letter was recommended for publication by Editor J. Civera upon evaluation of Reviewers' comments. (Corresponding author: Feng Zhu.)}
	\thanks{		
		Yunge Cui and Feng Zhu are with the Key Laboratory of Opto-Electronic Information Processing, Shenyang Institute of Automation, Chinese Academy of Sciences, Shenyang, China, (e-mail: cuiyunge@sia.cn; fzhu@sia.cn). Yinlong Zhang is with the Key Laboratory of Networked Control Systems, Shenyang Institute of Automation, Chinese Academy of Sciences, Shenyang, China. Jiahua Dong is with the State Key Laboratory of Robotics, Shenyang Institute of Automation, Chinese Academy of Science, Shenyang, China. They are also with the Institutes for Robotics and Intelligent Manufacturing, Chinese Academy of
		Sciences, Shenyang 110016, China, and are also with the University of Chinese Academy of Sciences, Beijing 101408, China. 
		 
		 Haibo Sun is with the Shanghai Institute of Microsystem and Information Technology, Chinese Academy of Sciences, Shanghai, China. 
		 
		 Xieyuanli Chen is with the College of Intelligence Science and Technology, National University of Defense Technology, Changsha, China.

 		} 
	 
	}

\maketitle

\markboth{IEEE Robotics and Automation Letters. Preprint Version. Accepted December, 2023}{Cui \MakeLowercase{\textit{et al.}}: LinK3D: Linear Keypoints Representation for 3D LiDAR Point Cloud}

\begin{abstract}
Feature extraction and matching are the basic parts of many robotic vision tasks, such as 2D or 3D object detection, recognition, and registration. As is known, 2D feature extraction and matching have already achieved great success. Unfortunately, in the field of 3D, the current methods may fail to support the extensive application of 3D LiDAR sensors in robotic vision tasks due to their poor descriptiveness and inefficiency. To address this limitation, we propose a novel 3D feature representation method: \underline{Lin}ear \underline{K}eypoints representation for \underline{3D} LiDAR point cloud, called LinK3D. The novelty of LinK3D lies in that it fully considers the characteristics (such as the sparsity and complexity) of LiDAR point clouds and represents the keypoint with its robust neighbor keypoints, which provide strong constraints in the description of the keypoint. The proposed LinK3D has been evaluated on three public datasets, and the experimental results show that our method achieves great matching performance. More importantly, LinK3D also shows excellent real-time performance, faster than the sensor frame rate at 10 Hz of a typical rotating LiDAR sensor. LinK3D only takes an average of $\bm{30}$ milliseconds to extract features from the point cloud collected by a 64-beam LiDAR and takes merely about $\bm{20}$ milliseconds to match two LiDAR scans when executed on a computer with an Intel Core i7 processor. Moreover, our method can be extended to LiDAR odometry task, and shows good scalability. We release the implementation of our method at \href{https://github.com/YungeCui/LinK3D}{https://github.com/YungeCui/LinK3D}.
\end{abstract}


\section{Introduction}
\IEEEPARstart{F}{eature} extraction and matching are the building blocks for most robotic vision tasks, such as object detection \cite{10044977}, and reconstruction \cite{9913658} tasks. In the field of 2D vision, a variety of famous 2D feature extraction methods (such as SIFT \cite{Lowe2004} and ORB \cite{Rublee2011}), have been proposed and widely used. However, in the field of 3D vision, there are still several unsolved issues for 3D feature representation and matching. Current methods \cite{Rusu2009, salti2014shot, prakhya2015b, lu2019deepvcp, du2020dh3d, bai2020d3feat, fischer2021stickypillars} may not be suitable for the high frequency (usually $\geq$ 10Hz) of 3D LiDAR and the large-scale complex scenes, especially in terms of efficiency and reliability. The irregularity, sparsity, and disorder of the LiDAR point cloud make it infeasible for 2D methods directly applied to 3D.

\begin{figure}[t]
	\centering
	\includegraphics[width=0.95\linewidth]{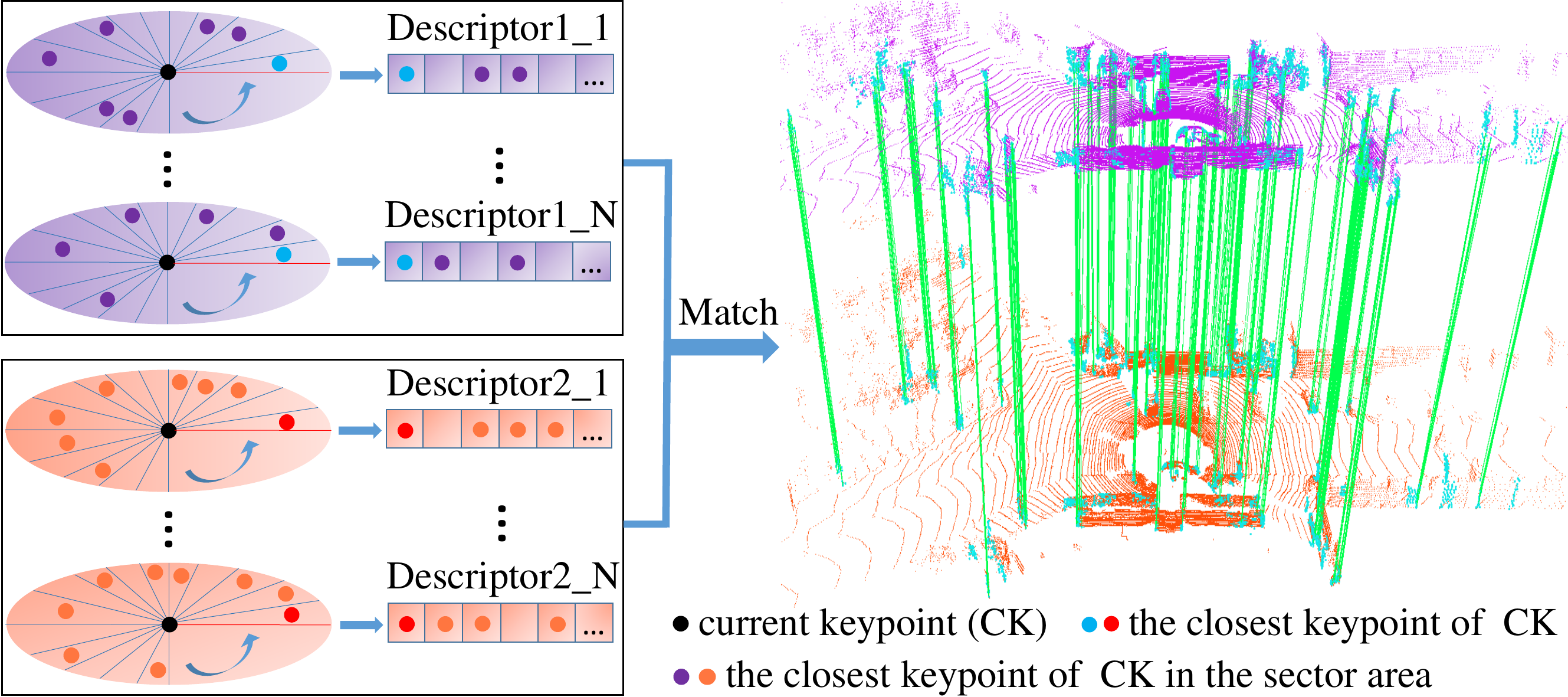}
	\caption{The core idea of the proposed LinK3D and the matching results of two LiDAR scans based on LinK3D. LinK3D represents the current keypoint with its neighbor keypoints. The green lines are the valid matches. Accurate point-to-point matches can be obtained by matching the corresponding LinK3D descriptors.}
	\label{fig:onecol}
	\label{fig1}
\end{figure}

Existing 3D feature point representation methods can be mainly divided into two categories in terms of extraction strategies, i.e., hand-crafted features and learning-based features. The hand-crafted features \cite{Rusu2008, Rusu2009, salti2014shot, prakhya2015b} mainly describe features in the form of histograms, and they use local or global statistical information to represent features. As there are usually many similar local features (such as local planes) in large-scale scenes that LiDAR uses, these local statistical features can easily lead to mismatches. The global features \cite{kim2021scan, fan2020seed}, intuitively, are unlikely to generate accurate point-to-point matches inside the point cloud. Learning-based methods \cite{yew20183dfeat, choy2019fully, lu2019deepvcp, du2020dh3d, bai2020d3feat, fischer2021stickypillars} have made great progress. However, the efficiency and generalization performance of these methods are still to be improved. In addition, some methods \cite{Rusu2009, salti2014shot, prakhya2015b} were proposed for the point clouds collected from small-scale object surfaces (e.g., the Stanford Bunny\footnotemark[1]\footnotetext[1]{http://graphics.stanford.edu/data/3Dscanrep/} point clouds). Obviously, there are some differences between the small-scale objects and the large-scale scenes using 3D LiDAR (e.g., the city scene of KITTI \cite{Geiger2012}). Specifically, the main differences are as follows: 
\begin{itemize}
	\item The small object's surface is usually smooth and continuous, and its local surface is unique. However, the 3D LiDAR point cloud contains lots of discontinuous and similar local surfaces (e.g., similar local planes, trees, poles, etc.), and they easily lead to mismatches.
	\item Compared with the point cloud of small-scale objects, the LiDAR point cloud is usually sparser, and the points are unevenly distributed in space. If there are not enough points in a fixed-size space, an effective statistical description may not be yielded. 
	\item Different from static and complete small-scale objects, there are often dynamic objects (cars, pedestrians, etc.) and occlusions in LiDAR scans. This can easily lead to inconsistent descriptions of the same local surface in the current and subsequent LiDAR scans.  
\end{itemize}

\begin{figure*}[t]
	\centering
	\includegraphics[width=0.85\linewidth]{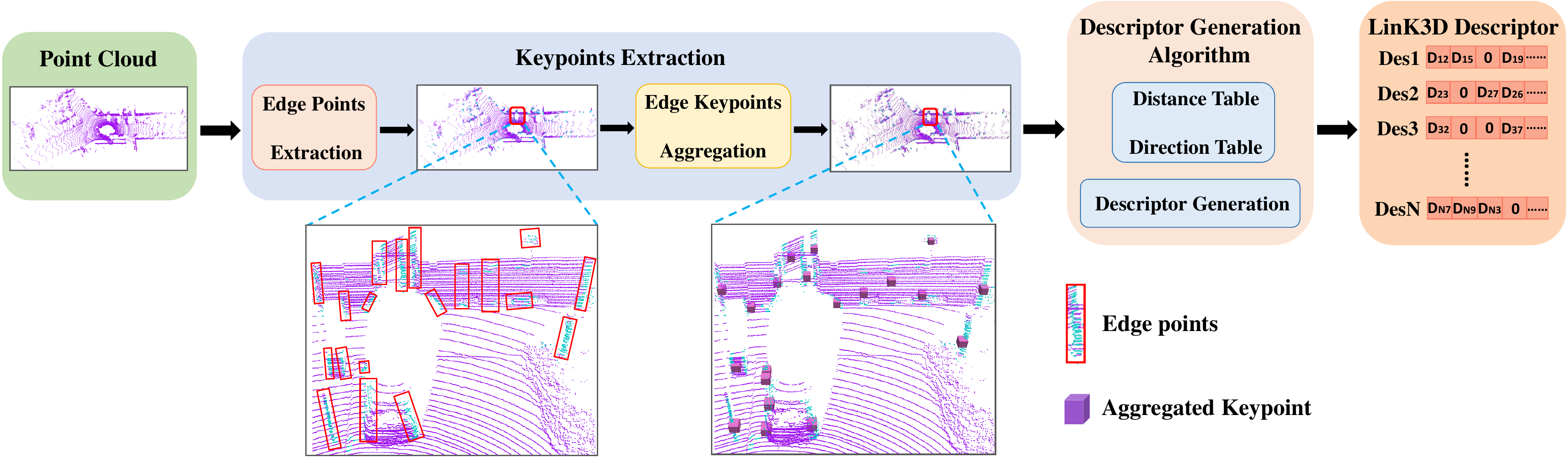}
	\caption{The workflow of the proposed LinK3D in terms of keypoint extraction and description. The keypoint extraction is first executed to generate aggregation keypoints. Afterward, the descriptor generation algorithm is performed to derive an efficient keypoint descriptor.}
	\label{fig2}
\end{figure*}

According to the differences, in this paper, we propose a novel 3D feature for LiDAR point clouds. Our method first extracts the robust aggregation keypoints. Then the extracted aggregation keypoints are fed into the descriptor generation algorithm. As shown in \cref{fig2}, the algorithm generates the descriptor in the form of a novel keypoint representation. After obtaining LinK3D descriptors, the matching algorithm can quickly match the descriptors of two LiDAR scans. In experiments, the proposed LinK3D achieves great matching performance and also shows impressive efficiency. Furthermore, LinK3D can be potentially applied to downstream 3D vision tasks, and we have applied LinK3D to LiDAR odometry. To summarize, our main contributions are as follows:\par
\begin{itemize}
	\item \textbf{Strong matching performance.} The proposed LinK3D feature considers the characteristics of LiDAR point clouds, and achieves significant progress in matching performance for sparse LiDAR point clouds.
	\item \textbf{Real-time performance on CPU.} The proposed LinK3D shows impressive efficiency, which makes it more suitable for the 3D applications of mobile robots with limited computing resources.
	\item \textbf{Good scalability.} LinK3D can be potentially applied to downstream 3D vision tasks. In this paper, LinK3D has been applied to the LiDAR odometry task.
\end{itemize} 

\section{Related Work} \label{Related Work}
Based on the extraction strategy, current 3D feature extraction methods can be divided into hand-crafted methods and deep neural network (DNN) methods. 

\textbf{Hand-crafted methods}. 
The histograms are usually used to represent different characteristics of the local surface. 
PFH \cite{Rusu2008} generates a multi-dimensional histogram feature of point pairs in the support region. FPFH \cite{Rusu2009} builds a Simplified Point Feature Histogram (SPFH) for each point by calculating the relationships between the point and its neighbors. SHOT \cite{salti2014shot} combines the spatial and geometric statical information and encodes the histograms of the surface normals in different spatial locations. In order to improve the matching efficiency, a binary quantization method B-SHOT \cite{prakhya2015b} is proposed that converts a real-valued vector to a binary vector. 3DHoPD \cite{Prakhya2017a} transforms the 3D keypoints into a new 3D space to generate histogram descriptions. In addition, the global descriptor Seed \cite{fan2020seed} is a segmentation-based method for the place recognition task of LiDAR SLAM. Moreover, GOSMatch \cite{zhu2020gosmatch} extracts the histogram-based graph descriptor for the place recognition of LiDAR SLAM. Due to the sparsity of the LiDAR point cloud, the statistical methods may fail to generate effective feature representation when there are not enough points.

\textbf{DNN-based methods.} 3DFeatNet \cite{yew20183dfeat} learns both 3D feature detectors and descriptors for point cloud matching using weak supervision. FCGF \cite{choy2019fully} extracts 3D features in a single pass by a 3D fully-convolutional network and presents metric learning losses to improve performance. DeepVCP \cite{lu2019deepvcp}  generates keypoints based on learned matching probabilities among a group of candidates. DH3D \cite{du2020dh3d} designs a hierarchical network to perform local feature detection, local feature description, and global descriptor extraction in a single forward pass. D3Feat \cite{bai2020d3feat} utilizes a self-supervised detector loss guided by the on-the-fly feature matching results during training. The semantic graph representation method \cite{kong2020semantic} reserves the semantic and topological information of the raw point cloud for the place recognition of LiDAR SLAM. StickyPillars \cite{fischer2021stickypillars} uses a handcrafted method to extract keypoints and combines the DNN method to generate descriptors, which is efficient in keypoint extraction but inefficient in descriptor generation. Geo Transformer \cite{qin2022geometric} encodes pair-wise distances and triplet-wise angles, making it robust in low-overlap cases and invariant to rigid transformation. In general, DNN-based methods usually require GPUs to speed up processing. In addition, the generalization of these methods is yet to be improved.

\section{Methodology} \label{The Proposed Method}
The pipeline of our method mainly consists of two parts: feature extraction and feature matching. The process of feature extraction is shown in \cref{fig1}. The edge points of LiDAR scans are first extracted, then they are fed into the edge keypoint aggregation algorithm, where the robust aggregation keypoints are further extracted for subsequent descriptor generation. In the descriptor generation algorithm, the distance table and the direction table are built for fast descriptor generation.

\begin{figure}
	\centering    
	\includegraphics[width=0.9\linewidth]{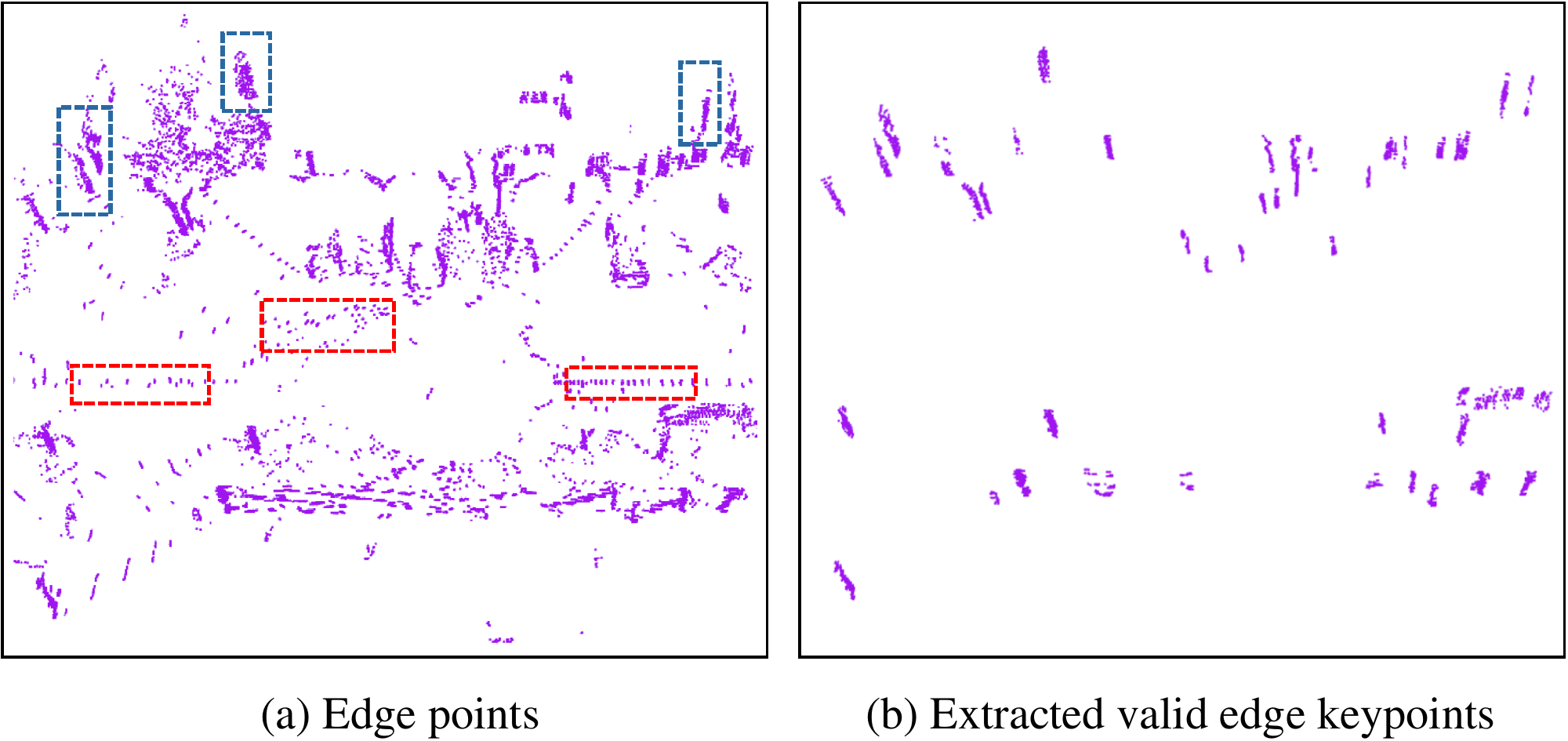}
	\caption{The scattered edge points (marked by the red dashed box) and the clustered edge keypoints (marked by the blue dashed box) are in (a). The clustered edge keypoints are what we need. (b) shows the extracted edge keypoints by Algorithm \ref{alg:A}.}
	\label{fig3}
\end{figure}

\subsection{Keypoint Extraction}
\subsubsection{Edge Point Extraction} \label{Keypoints Extraction Module}
In keypoint extraction, we roughly divide a LiDAR point cloud into two types: edge points and plane points. The main difference between edge points and plane points is the smoothness of the local surface where the points are located. Given a 3D LiDAR point cloud $P_c$, let $i$ be a point in $P_c$. $P_s$ is a set of continuous points on the same scan line as point $i$, and evenly distributed on both sides of $i$. $|S|$ is the cardinality of $P_s$. The smooth term of the current point $i$ is defined as follows:
\begin{equation}
	\label{equation1}
	\begin{aligned}
		\nabla_i = \frac{1}{|S|} {\Vert \sum\limits_{j \in P_s , j \neq i} \left(\vec{p_j} - \vec{p_i} \right)\Vert}^2
	\end{aligned}
\end{equation}
where $\vec{p_i}$ and $\vec{p_j}$ are the coordinates of the two points $i$ and $j$, respectively. The edge points (as shown in \cref{fig3}a) are extracted with $\nabla$ greater than a threshold $Th_{\nabla}$. \par

\subsubsection{Edge Keypoint Aggregation}
After obtaining the edge points, there are lots of points whose $\nabla$ are above the threshold $Th_{\nabla}$, but they are not stable. Specifically, these unstable points appear in the current scan but may not appear in the next scan. As marked by the red dashed boxes in \cref{fig3}a, the unstable points are usually scattered. Therefore, it is necessary to filter out these points and find the valid edge keypoints. As marked by the blue dashed boxes in \cref{fig3}a, the valid edge keypoints are usually distributed vertically in clusters.\par

\begin{algorithm}[h!]
	\IncMargin{1em}
	\caption{Keypoints Aggregation Algorithm}
	\label{alg:A}
	\SetKwInOut{Input}{\textbf{Input}}\SetKwInOut{Output}{\textbf{Output}}
	
	\Input{
		\textbf{$P_e$}: $\nabla \ \textgreater \ Th_{\nabla}$ edge points}
	\Output{$ValidEdgeKeypoints$, $AggregationKeypoints$}	
	\textbf{Main Loop:}
	\\
	\For{each point $p_i \in P_e $}{
		 Sectors$\leftarrow$DividePointToSectorBasedOnEq.\ref{equation2}($p_i$)\;
	}
	\For{each $Sector \in Sectors$ }{
		$FirstCluster$ $\leftarrow$ CreateCluster(any $point$ in $Sector$)\;
		$Clusters$.InsertCluster($FirstCluster$)\;
		\For{other $point$ $p_j$ $\in$ $Sector$ }{
			\For{each $Cluster \in Clusters$}{
				$Center$ $\leftarrow$ ComputeClusterCenter($Cluster$)\;
				$dist$ $\leftarrow$ ComputeHorizontalDist($p_j$, $Center$)\;
				\uIf{dist $\textless$ $Th_{dist}$}{
					$Cluster$.UpdateCluster($p_j$)\;
				}
				\uElseIf{$Cluster$ is the end one}{
					$NewCluster$ $\leftarrow$ CreateCluster($p_j$)\;
					$Clusters$.Insert($NewCluster$)\;
				}\Else{continue;}
			}
		}
		
		\For{each Cluster $\in Clusters$}{
			$N_{point}$ $\leftarrow$ CountNumberOfPoint($Cluster$)\;
			$N_{line}$ $\leftarrow$ CountNumberOfScanLine($Cluster$)\;
			\If{$N_{point} \ \textgreater \ Th_{point} \ \&\& \ N_{line}  \ \textgreater \ Th_{line}$}{
				$ValidEdgeKeypoints$.Insert($Cluster.Points$)\;			
				 $AggregationKeypoints$.Insert($Cluster.Center$);
			}
		}
	}
\end{algorithm}

\begin{figure}[t]
	\centering
	\includegraphics[width=0.65\linewidth]{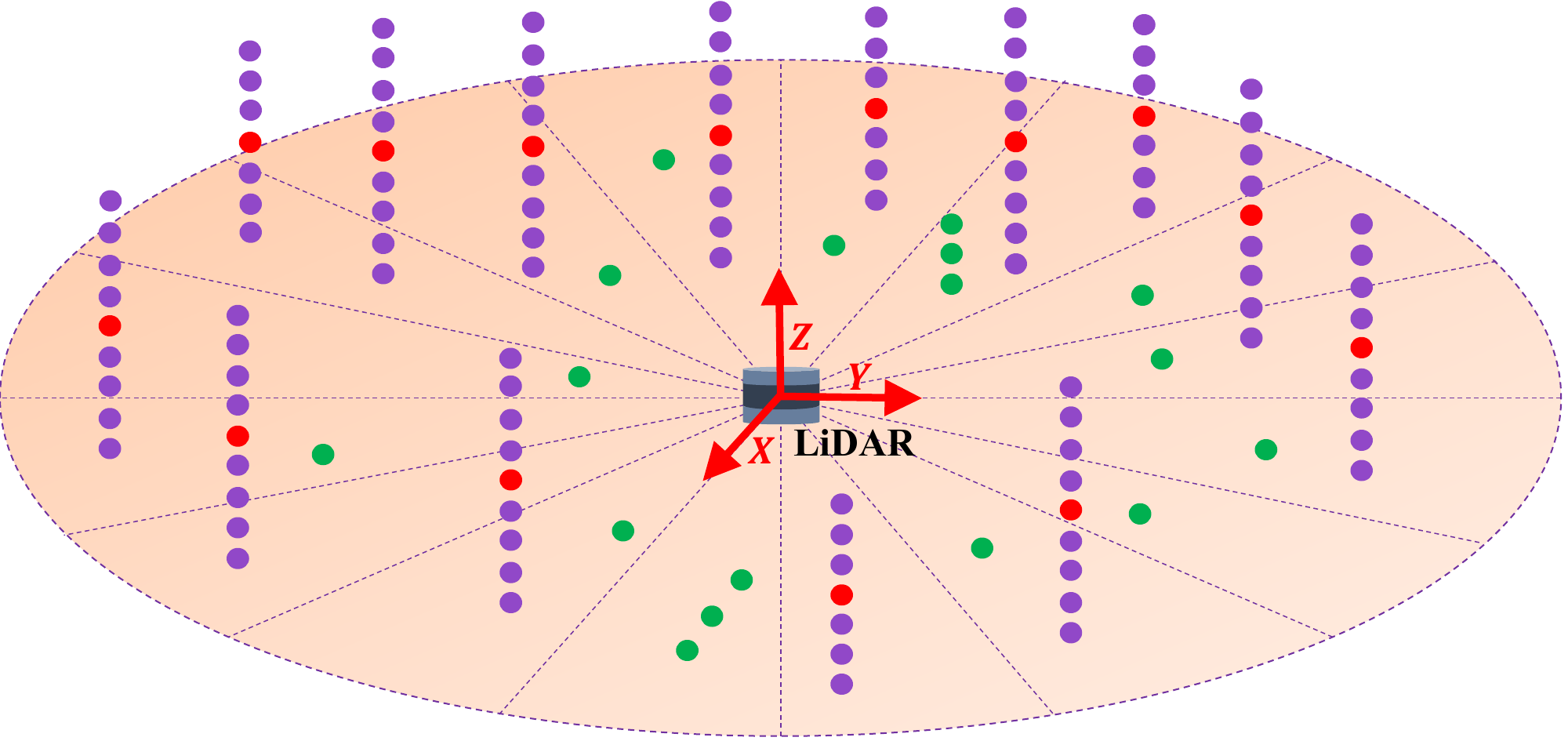}	
	\caption{Illustration of the aggregation process. The points are first divided into corresponding sector areas based on the angle in the XoY plane of the LiDAR coordinate system. Then the clustering operation is only performed within each sector area, rather than directly within the whole space. The purple points are what we need to generate the aggregation keypoints (red points), and the green points are the scattered points that will be filtered out.}
	\label{fig:onecol}
	\label{figAggregation}
\end{figure}

In this paper, a keypoint aggregation algorithm is designed to find valid edge keypoints. As illustrated in Fig.\ref{figAggregation}, the angle information guidance is used for accelerating the aggregation process. The motivation is that the points belonging to the same vertical edge usually have approximately the same angle in the XoY plane (assuming the Z-axis is upward) of the LiDAR coordinate system. The angle of point $\vec{p_i}$ is given by:
\begin{equation}
	\label{equation2}
	\begin{aligned}
		\theta_i = \arctan(\vec{p_i}.y/\vec{p_i}.x)
	\end{aligned}
\end{equation}
Therefore, we first divide the XoY plane centered on the origin of the LiDAR coordinate system into $N_{sect}$ sector areas equally, then only cluster the points in each sector area rather than cluster in the whole space.

The specific algorithm is shown in Algorithm \ref{alg:A}. It is worth noting that our algorithm runs about 25 times faster than the classical K-Means algorithm when we set $N_{sect} = 120$ in our experiment. The extracted edge keypoints are shown in \cref{fig3}b. It can be seen that our algorithm can filter out the invalid edge points and find the positive edge keypoints. In addition, the centroid of each cluster point is calculated and named as an aggregation keypoint, which will be used for subsequent descriptor generation. \par

\begin{figure}[t]
	\begin{center}
		\includegraphics[width=0.55\linewidth]{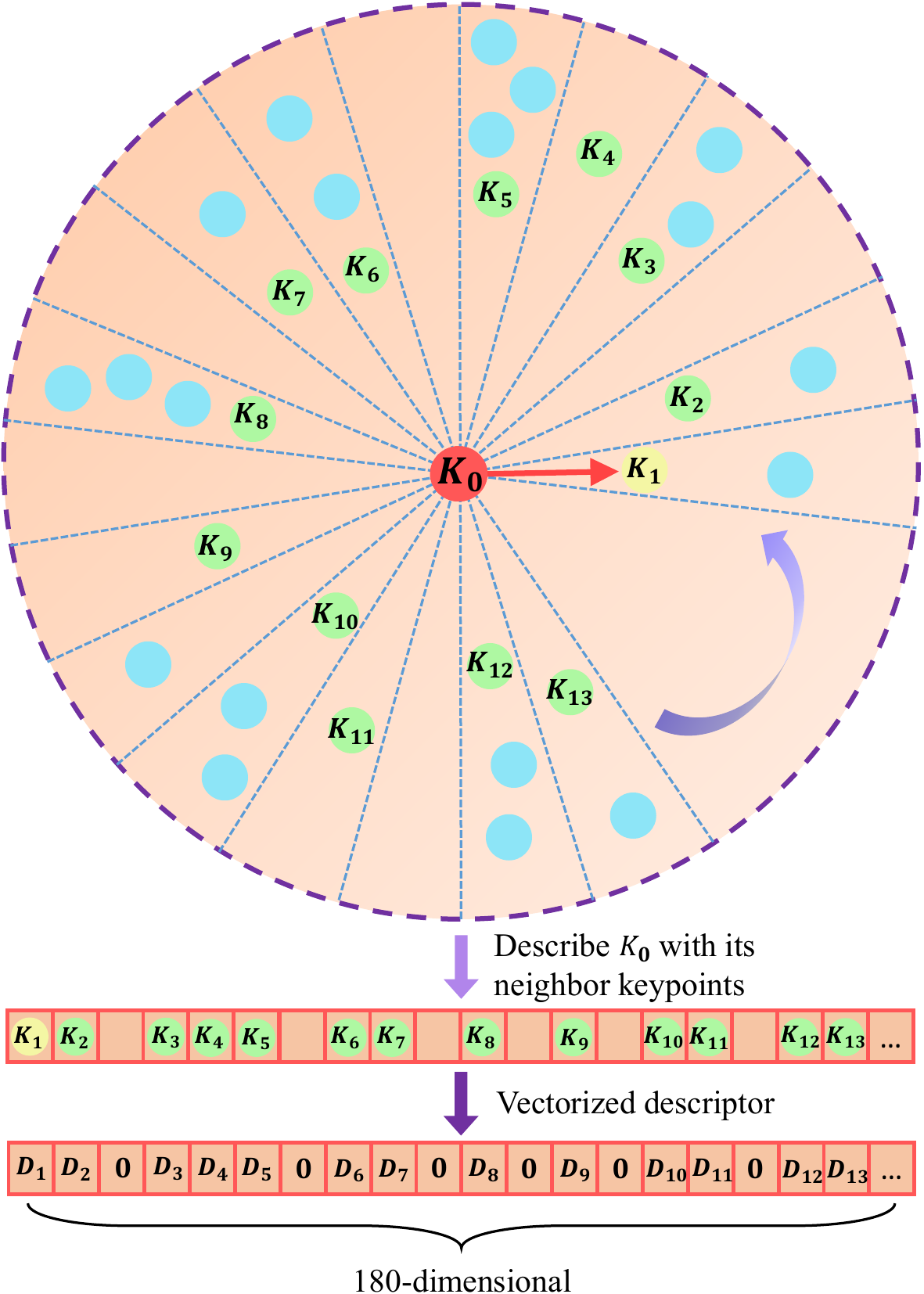}
	\end{center}
	\caption{Illustration of the descriptor generation. The XoY plane, centered on the current keypoint $k_o$ is divided into 180 sector areas. We first search for the closest keypoint $k_1$ of $k_0$, then the main direction is the vector from $k_0$ to $k_1$. Then the closest keypoint in each sector area is searched for. The searched keypoints $k_1$, $k_2$, $k_3$, etc., are used for describing the current $k_0$. Through the vectorizing operation, by using the distance values between $k_o$ and $k_1$, $k_2$, $k_3$, etc., the 180-dimensional LinK3D descriptor will be obtained.}
	\label{fig:long}
	\label{fig:onecol}
	\label{figDesgeneration}
\end{figure}

\subsection{Descriptor Generation} \label{Descriptor Generation}
\subsubsection{Descriptor Generation Process}
In descriptor generation, all aggregation keypoints are first projected to the XoY plane (assuming the Z-axis of LiDAR is upward), which can eliminate the influence caused by the uneven distribution of clustered edge keypoints along the Z-axis direction.
For fast matching, our LinK3D descriptor is represented as a 180-dimensional vector, which uses 0 or the distance between the current keypoint and its neighboring keypoints to represent each dimension. As shown in \cref{figDesgeneration}, we divide the XoY plane into 180 sector areas centered on the current keypoint $k_0$, and each descriptor dimension corresponds to a sector area. Inspired by the 2D descriptor SIFT \cite{Lowe2004}, which searches the main direction to ensure the rotation invariance, the main direction of LinK3D is also searched and represented as the direction vector from the current keypoint $k_0$ to its closest keypoint $k_1$, which is located in the first sector area. The other sector areas are arranged in counterclockwise order. Afterward, the closest keypoint of $k_0$ is searched in each sector. If there is the closest keypoint in a sector, we use the distance between the current keypoint and the closest keypoint to represent the corresponding dimension value in the descriptor. Otherwise, the value is set to 0.\par

During the process, the direction from the current point $k_0$ to other points $k_j(j \neq 1)$ is expressed as $\vec{m}_{0j}$, and we use the angle between ${\vec m_{0j}}$ and the main direction $\vec{m}_{01}$ to determine which sector $k_j$ belongs to. The angle is calculated by:
\begin{equation} 
	\label{direction}
	\begin{aligned}
		\theta_j=
		\begin{cases}
			\arccos \frac{\vec{m}_{01} \cdot \vec{m}_{0j}}{\vert \vec{m}_{01} \vert \vert \vec{m}_{0j} \vert }   & if \ D_j>0 \\
			2\pi - \arccos\frac{\vec{m}_{01} \cdot \vec{m}_{0j}}{\vert \vec{m}_{01} \vert \vert \vec{m}_{0j} \vert }   & if \ D_j<0
		\end{cases}
	\end{aligned}
\end{equation}
where $D_j$ is defined as:
\begin{equation}
	D_i = \left| \begin{array}{cc}
		x_1 & y_1 \\
		x_j & y_j
	\end{array} \right|
\end{equation}

\begin{algorithm}[htb]
	\IncMargin{1em}
	\caption{Descriptor Generation Algorithm}
	\label{alg:B}
	\SetKwInOut{Input}{\textbf{Input}}\SetKwInOut{Output}{\textbf{Output}}
	\Input{
		$K_a$: AggregationKeypoints
	}
	\Output{
		$Descriptors$
	}
	\textbf{Main Loop:}
	\\
	\For{each point $k_i \in K_a$}{
		\For{each point $k_j \in K_a, k_j \neq k_i$}{
			$Table_{dist}$ $\leftarrow$ ComputeDistance($k_i$, $k_j$)\;	
			$Table_{dire}$ $\leftarrow$ ComputeDirection($k_i$, $k_j$)\;
		}
	}
	\For{each point $k_i \in K_a$}{
		$ClosestPt$ $\leftarrow$ SearchClosestPoint($k_i$ , $Table_{dist}$)\;
		$Sectors$.InsertPointToFirstSector($ClosestPt$) \;
		$MainDire \leftarrow$ GetDirection($k_i, ClosestPt, Table_{dire}$)\;
		\For{each point $k_j \in K_a , k_j \neq k_i, k_j \neq ClosestPoint $}{
			$OtherDire \leftarrow$ GetDirection($k_i , k_j, Table_{dire}$)\;
			$\theta_j$ $\leftarrow$ Compute$\theta$InEq\ref{direction}($MainDire, OtherDire$)\;	
			$Sectors$.InsertPointBasedOn$\theta$($k_j$, $\theta_j$)\;	
		}
		Define a 180-dimensional $Descriptor$\;
		\For{each sector $\in Sectors$}{			
			\uIf{$sector.NumberOfPoints == 0 $}{
				$CorrespondingDimentionInDescriptor = 0$\;
			}
			\Else{
				$Dis \leftarrow$ SearchClosestDist($sector$, $Table_{dist}$)\;
				$CorrespondingDimInDescriptor =  Dis$\;
			}
		}
		$Descriptors$.InsertNewDescriptor($Descriptor$)\;
		
	}
\end{algorithm}

\subsubsection{Issues of Descriptor Generation Process}
\begin{figure}[htb]
	\begin{center}
		\includegraphics[width=0.6\linewidth]{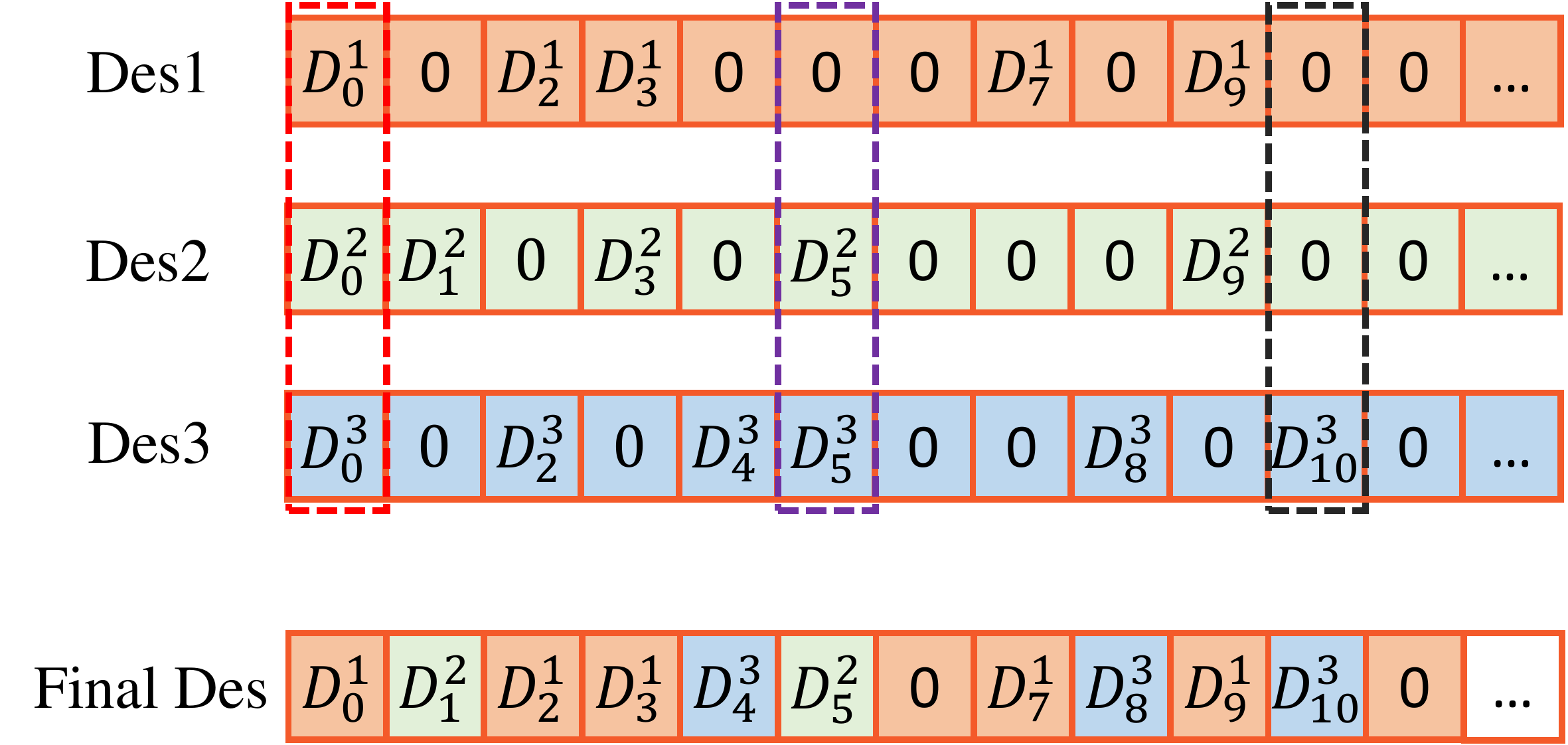}
	\end{center}
	\caption{The value of each dimension in the final descriptor corresponds to the non-zero value with the highest priority among Des1, Des2, and Des3.}
	\label{fig:long}
	\label{fig:onecol}
	\label{figPriority}
\end{figure}

There are two main issues with the above-mentioned algorithm. One issue is that the algorithm is sensitive to the closest keypoint. In the presence of interference from an outlier keypoint, the matching will fail. The other issue is that we have to calculate the relative distance and direction between two points frequently, so there will be lots of repeated calculations. To solve the first issue, we search for a certain number of the closest keypoints. Suppose we search for the 3 closest keypoints, and the corresponding 3 descriptors are calculated, as shown in \cref{figPriority}. Des1 corresponds to the closest keypoint, and Des3 corresponds to the third closest keypoint. We define the priorities based on the distances between them and the current keypoint. Des1 has the highest priority due to its closest distance, and Des3 has the lowest priority because of its furthest distance. The value of each dimension in the final descriptor corresponds to the non-zero value with the highest priority among them. As marked by the red dashed box in \cref{figPriority}, Des1 has a non-zero value $D_0^1$, and its corresponding value in the final descriptor is also set as $D_0^1$ due to its high priority. The other two cases are shown in purple and black dashed boxes in \cref{figPriority}. To solve the second issue, we build the distance table and the direction table for all keypoints to avoid repeated calculations by looking up the tables. The specific descriptor generation is shown in Algorithm \ref{alg:B}, which shows the process of extracting one descriptor.\par

\begin{algorithm}[htb]
	\IncMargin{1em}
	\caption{Matching Algorithm}
	\label{alg:C}
	\SetKwInOut{Input}{\textbf{Input}}\SetKwInOut{Output}{\textbf{Output}}
	
	\Input{
		\textbf{$Descriptors$_$1$, $Descriptors$_$2$} }
	\Output{
		$MatchPairs$: matched descriptor index pairs\\
	}	
	\textbf{Main Loop:}
	\\
	Define $SetOfID_j$\;
	Define $RBtree$_$ID_j$_$ID_i$, $RBtree$_$ID_i$_$Score$\;
	\For{each descriptor $D_i \in$ $Descriptors$_$1$ }{
		Define $HighestScore$ and $HighestScoreID_j$\;
		\For{each descriptor $D_j \in$ $Descriptors$_$2$}{
			$Score$ $\leftarrow$ GetSimilarityScore($D_i, D_j$)\;
			\If{$Score$ $\textgreater$ $HighestScore$}{
				$HighestScore$ = $Score$\;
				$HighestScoreID_j$ = $ID_j$\;
			}
		}
		$SetOfID_j$.Insert($HighestScoreID_j$)\;
		$RBtree$_$ID_j$_$ID_i$.Insert($HighestScoreID_j$, $ID_i$)\;
		$RBtree$_$ID_i$_$Score$.Insert($ID_i$, $HighestScore$)\;
		
	}
	
	\begin{tikzpicture}
		\node[draw, densely dashed] {$\#$ Removing one-to-multiple matches for $Descriptors$_$2$ $\#$}\;
	\end{tikzpicture}
	
	\For{each $ID_j$ $\in$ $SetOfID_j$}{		
		$AllID_i$ $\leftarrow$ GetAllOf$D_i$($ID_j$, $RBtree$_$ID_j$_$ID_i$)\;
		
		$HighestScore$_$ID_i$ $\leftarrow$ GetHighestScore$D_i$($AllID_i$, $RBtree$_$ID_i$_$Score$)\;
		
		\If{ $HighestScore$_$ID_i$.$Score$ $\geq Th_{score}$ }{
			$MatchPairs$.Insert($HighestScore$_$ID_i$.$ID_i$, $ID_j$)\;
		}
	}
\end{algorithm}

\subsection{Matching Two Descriptors} \label{Similarity Measurement and Reverse Matching}
To measure the similarity of two descriptors, we only count the non-zero dimensions of two descriptors. Specifically, the absolute value of the difference between the corresponding non-zero dimension is computed in the two descriptors. If the difference is less than 0.2, the similarity score is increased by 1. If the similarity score of a match is greater than the threshold $Th_{score}$, the match is considered valid. The specific matching algorithm is shown in Algorithm \ref{alg:C}. After matching two descriptors, the matches between the edge keypoints are searched in clusters. Specifically, the corresponding edge keypoints that have the highest $\nabla$ (smooth term) in each scan line are first selected. Then, the edge keypoints located on the same scan line are matched.\par

\section{Experiments} \label{experiments}
In this section, we first perform the basic matching performance and real-time performance evaluation, then we extend LinK3D to the LiDAR odometry task. KITTI odometry \cite{Geiger2012}, M2DGR \cite{yin2021m2dgr} and StevenVLP \cite{Shan2018} datasets are used for the evaluation. The point clouds in KITTI were collected from different street environments (such as inner city, suburb, forest, and high-way) by a 64-beam Velodyne LiDAR at a rate of 10 Hz. The point clouds of M2DGR are collected by a 32-beam Velodyne LiDAR, and the point clouds of Steven were collected from the campus scene by a 16-beam Velodyne LiDAR at a rate of 10 Hz. The frequency (10 Hz) of the LiDAR requires the processing time of all algorithms to run within 100 milliseconds. The parameters of our algorithm are set as follows: $Th_{\nabla} = 10$ in \cref{Keypoints Extraction Module};  $Th_{dist} = 0.4$, $Th_{point} = 12$, $Th_{line} = 4$ in Algorithm \ref{alg:A}; $Th_{score} = 3$ in \cref{Similarity Measurement and Reverse Matching}. The code is executed on a desktop with an Intel Core i7-12700 @ 2.10 GHz processor and 16 GB of RAM.

\begin{figure*}
	\centering    
	\includegraphics[width=0.95\linewidth]{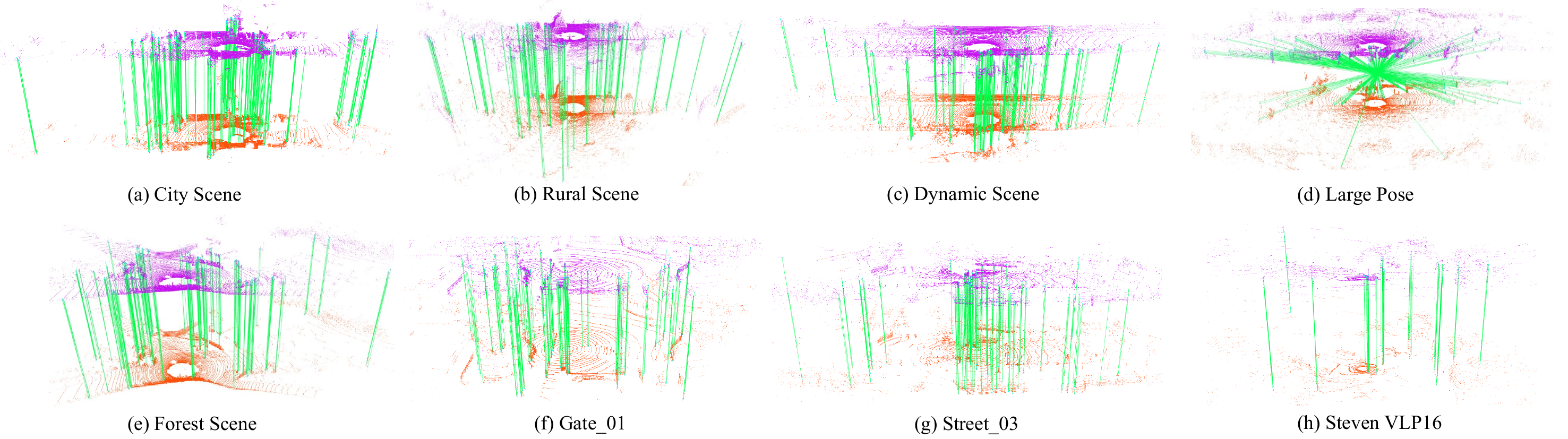}
	\caption{The matching results of LinK3D on different scenes of KITTI, M2DGR, and Steven VLP16. The green lines are the valid matches after RANSAC.}
	\label{fig6}
\end{figure*}

\begin{table*}[htb]
	\centering
	\caption{The number of extracted keypoints in different scenes. The specific IDs of two matched scans are enclosed in parentheses.}
	\begin{tabular}{ c| c c c c c c c c }
		\toprule
		Keypoint & \makecell[c]{City Scene\\ (169/170)} & \makecell[c]{Rural Scene \\ (580/581)} & \makecell[c]{Dynamic Scene\\ (87/88)} & \makecell[c]{Lage Pose \\ (232/1647)} & \makecell[c]{Forest Scene \\ (17/18)} & \makecell[c]{Gate_01 \\ (121/122)} & \makecell[c]{Street_03 \\ (109/110)} & \makecell[c]{Steven VLP16 \\ (22/23)} \\
		\midrule
		ISS \cite{Zhong2009} & 2985/2863 & 2459/2508 & 3460/3396 &  2467/2404 & 2869/2949 & 1184/1143 & 1153/1120 & 222/213\\
		Sift3D \cite{scovanner20073} & 2897/2842 & 2485/2442 & 3428/3356 & 2497/2478 & 2853/2877 & 1162/1155 & 1248/1158 & 229/237\\
		LinK3D(ours) & 2890/2902 & 2359/2284 & 3380/3428 & 2411/2604 & 2928/3006 & 1177/1162 & 1204/1156 & 218/226\\
		\bottomrule
	\end{tabular}
	\label{tab:Num}
\end{table*}

\begin{table*} [!htbp]
	\centering
	\caption{The comparison results of matching performance in different scenes of the KITTI dataset.}
	\begin{tabular}{c c | c  c  c | c  c  c | c  c  c | c  c  c}
		\toprule
		\multirow{2}{*}{Keypoint} &\multirow{2}{*}{Descriptor}  & \multicolumn{3}{c|}{ City Scene} & \multicolumn{3}{c|}{ Rural Scene} & \multicolumn{3}{c|}{ Dynamic Scene} & \multicolumn{3}{c}{Large Pose} \\
		\cmidrule{3-14}
		& & $N_{M}$ & $N_{inliers}$ & $inliers\%$ & $N_{M}$& $N_{inliers}$ & $inliers\%$ & $N_{M}$ & $N_{inliers}$ & $inliers\%$  & $N_{M}$ & $N_{inliers}$ & $inliers\%$\\
		\midrule
		\multirow{6}{*}{LinK3D} &3DHoPD\cite{Prakhya2017a}  & 32 & 0 & 0 & 20 & 0 & 0 & 49 & 0 & 0 & 35 & 0 & 0 \\
		&PFH \cite{Rusu2008} & 60 & 11 & 18.3 & 17 & 5 & 29.4 & 65 & 0 & 0 & 31 & 0 & 0 \\
		&3DSC\cite{Frome2004}  & 1 & 0 & 0 & 0 & 0 & - & 1 & 0 & 0 & 0 & 0 & - \\
		&SHOT\cite{Tombari2010}  & 0 & 0 & - & 0 & 0 & - & 2 & 0 & 0 & 0 & 0 & - \\
		&FPFH \cite{Rusu2009} & 8 & 0 & 0 & 0 & 0 & - & 6 & 0 & 0 & 1 & 0 & 0 \\
		&BSHOT\cite{prakhya2015b} & 45 & 33 & \textbf{73.3} & 20 & 7 & 35.0 & 35 & 15 & \textbf{42.9} & 24 & 12 & \textbf{50.0} \\
		\midrule
		\multicolumn{2}{c|}{LinK3D (Aggregation)} & $135$  & \textbf{71} & 52.6 & $110$ & \textbf{48} & \textbf{43.6}  & $147$ & \textbf{39} & 26.5 & $134$ & \textbf{50} & 37.3\\
		\midrule
		\midrule
		\multirow{6}{*}{ISS}&3DHoPD\cite{Prakhya2017a} & 2423 & 280 & 11.6 & 2119 & 83 & 3.9 & 2967 & 341 & 11.5 & 2001 & 7 & 0.3\\ 
		&PFH\cite{Rusu2008} & 1916 & 360 & 18.8 & 839 & 66 & 7.9  & 1943 & 75 & 3.9 & 717 & 9 & 1.3\\
		&3DSC\cite{Frome2004} & 1987 & 6 & 0.3 & 2214 & 4 & 0.2 & 2969 & 4 & 0.13 & 2169 & 0 & 0 \\
		&SHOT\cite{Tombari2010} & 219 & 111 & 50.7 & 89 & 34 & 38.2 & 479 & 121 & 25.2 & 23 & 4 & 17.4\\
		&FPFH \cite{Rusu2009} & 423 & 101 & 23.9 & 104 & 8 & 7.7 & 442 & 34 & 7.7 & 193 & 4 & 2.1\\
		&BSHOT\cite{prakhya2015b} & 641 & 380 & 59.3 & 330 & 109 & 33.0 & 539 & 129 & 23.9 & 274 & 52 & 19.0 \\
		\midrule
		\multirow{6}{*}{Sift3D} &3DHoPD\cite{Prakhya2017a} & 2197 & 162 & 7.4 & 1809 & 3 & 0.17 & 2736 & 224 & 8.2 & 1722 & 4 & 0.2\\
		&PFH \cite{Rusu2008} & 1803 & 383 & 21.2 & 743 & 5 & 0.7 & 1753 & 116 & 6.6 & 755 & 0 & 0\\
		&3DSC\cite{Frome2004} & 1777 & 0 & 0 & 1840 & 0 & 0 & 2336 & 0 & 0 & 1874 & 0 & 0\\
		&SHOT\cite{Tombari2010}  & 158 & 80 & 50.6 & 35 & 20 & \textbf{57.1} & 298 & 111 & 37.2 & 9 & 0 & 0\\
		&FPFH \cite{Rusu2009} & 534 & 113 & 21.2 & 238 & 0 & 0  & 458 & 25 & 5.5 & 340 & 0 & 0\\
		&BSHOT\cite{prakhya2015b}  & 623 & 403 & \textbf{64.7} & 332 & 72 & 21.7 & 495 & 137 & 27.7 & 246 & 47 & 19.1\\
		\midrule
		\multicolumn{2}{c|}{LinK3D (Edge)} & $1090$  & \textbf{602} & 55.2 & $816$ & \textbf{403} & 49.4 & $1068$ & \textbf{413} & \textbf{38.7}  & $681$ & \textbf{367} & \textbf{53.9} \\
		\bottomrule		
	\end{tabular}
	\label{tab:II}
\end{table*}

\begin{table*} [!htbp]
	\centering
	\caption{The comparison results of matching performance on the forest scene of KITTI and the M2DGR and Steven VLP16 datasets. }
	\begin{tabular}{c c | c  c  c | c  c  c | c  c  c | c  c  c }
		\toprule
		\multirow{2}{*}{Keypoint} &\multirow{2}{*}{Descriptor} & \multicolumn{3}{c|}{Forest Scene}  & \multicolumn{3}{c|}{Gate_01} & \multicolumn{3}{c|}{Street_03} & \multicolumn{3}{c}{StevenVLP} \\
		\cmidrule{3-14}
		& & $N_{M}$ & $N_{inliers}$ & $inliers\%$ & $N_{M}$ & $N_{inliers}$ & $inliers\%$ & $N_{M}$& $N_{inliers}$ & $inliers\%$ & $N_{M}$& $N_{inliers}$ & $inliers\%$ \\
		\midrule
		\multirow{6}{*}{LinK3D} &3DHoPD\cite{Prakhya2017a} & 31 & 0 & 0 & 17 & 7 & 41.2 & 23 & 0 & 0 & 1 & 0 & 0 \\
		&PFH \cite{Rusu2008} & 68 & 5 & 7.4 & 27 & 16 & 59.3 & 16 & 10 & 62.5 & 3 & 0 & 0 \\
		&3DSC\cite{Frome2004} & 0 & 0 & - & 0 & 0 & - & 0 & 0 & - & 0 & 0 & - \\
		&SHOT\cite{Tombari2010} & 0 & 0 & - & 0 & 0 & - & 2 & 0 & 0 & 1 & 0 & 0\\
		&FPFH \cite{Rusu2009} & 26 & 0 & 0 & 2 & 0 & 0 & 2 & 0 & 0 & 3 & 0 & 0 \\
		&BSHOT\cite{prakhya2015b} & 48 & 28 & \textbf{58.3} & 29 & 21 & 72.4 & 33 & 21 & 63.6 & 10 & 5 & 50.0\\
		\midrule
		\multicolumn{2}{c|}{LinK3D (Aggregation)} & 147 & \textbf{61} & 41.5 & $70$  & \textbf{56} & \textbf{80.0} & 77 & \textbf{54} & \textbf{70.1} & $15$ & \textbf{12} & \textbf{80.0} \\
		\midrule
		\midrule
		\multirow{6}{*}{ISS}&3DHoPD\cite{Prakhya2017a} & 2447 & 337 & 13.8 & 775 & 38 & 4.9 & 741 & 4 & 0.5 & 51 & 0 & 0\\ 
		&PFH\cite{Rusu2008} & 1507 & 98 & 6.5 & 617 & 4 & 0.6 & 540 & 0 & 0 & 141 & 0 & 0 \\
		&3DSC\cite{Frome2004} & 2530 & 4 & 0.16 & 1057 & 0 & 0 & 1004 & 0 & 0 & 203 & 0 & 0 \\
		&SHOT\cite{Tombari2010} & 159 & 88 & 55.3 & 30 & 19 & 63.3 & 22 & 15 & 68.2 & 0 & 0 & - \\
		&FPFH \cite{Rusu2009} & 340 & 32 & 9.4 & 476 & 0 & 0 & 393 & 0 & 0 & 120 & 0 & 0 \\
		&BSHOT\cite{prakhya2015b} & 438 & 228 & 52.1 & 253 & 92 & 36.4 & 253 & 113 & 44.7 & 59 & 10 & 16.9 \\
		\midrule
		\multirow{6}{*}{Sift3D} &3DHoPD\cite{Prakhya2017a} & 2128 & 195 & 9.2 & 673 & 0 & 0 & 669 & 6 & 0.9 & 82 & 0 & 0 \\
		&PFH \cite{Rusu2008} & 1310 & 100 & 7.6 & 636 & 0 & 0 & 518 & 0 & 0 & 142 & 0 & 0 \\
		&3DSC\cite{Frome2004} & 1679 & 0 & 0 & 851 & 0 & 0 & 866 & 0 & 0 & 116 & 0 & 0 \\
		&SHOT\cite{Tombari2010} & 82 & 56 & \textbf{68.3} & 19 & 12 & 63.2 & 17 & 16 & \textbf{94.1} & 1 & 0 & 0 \\
		&FPFH \cite{Rusu2009} & 343 & 5 & 1.5 & 429 & 0 & 0 & 337 & 0 & 0 & 115 & 0 & 0 \\
		&BSHOT\cite{prakhya2015b} & 387 & 192 & 49.6 & 236 & 102 & 43.2 & 218 & 97 & 44.5 & 25 & 9 & 36.0\\
		\midrule
		\multicolumn{2}{c|}{LinK3D (Edge)} & 1015 & \textbf{455} & 44.8 & $346$  & \textbf{280} & \textbf{80.9} & 405 & \textbf{298} & 73.6 & $100$ & \textbf{87} & \textbf{87.0} \\
		\bottomrule		
		
	\end{tabular}
	\label{tab:III}
\end{table*}

\subsection{Matching Performance Comparisons with Hand-Crafted Features} \label{matching performance}
Feature matching is a basic function for hand-crafted 3D local features. In this subsection, the typical city scene (Seq.00), rural scene (Seq.03), dynamic scene (Sec.04), large pose (Sec.08), and forest scene (Sec.09) in KITTI are used for the evaluation. Gate_01, Street_03 in M2DGR, and the first sequence in Steven VLP16 are used for the evaluation. The state-of-the-art handcrafted 3D features: 3DSC \cite{Frome2004}, PFH \cite{Rusu2008}, FPFH \cite{Rusu2009}, SHOT \cite{Tombari2010}, BSHOT \cite{prakhya2015b}, and 3DHoPD \cite{Prakhya2017a} are used for the comparison. We first extract our LinK3D aggregation keypoints for comparison methods and compare their matching results with our method on the same scenes. Then we also extract the ISS \cite{Zhong2009} and Sift3D \cite{scovanner20073} keypoints for comparison descriptors on the same scenes, and the number of the two keypoints is similar to the number of edge keypoints in LinK3D. For metric methods, we follow the metrics used in \cite{Rublee2011}, and use the number of inliers and $inliers \%$ as the validation metrics. RANSAC \cite{Fischler1981} is used for removing the mismatches, and its acceptance threshold is set as 0.5. For a fair comparison, we first compute the average number of inliers and $inliers \%$ for our method in different scenes (except the large pose situation, which is a randomly selected reverse loop), then use two matched LiDAR scans, the number of inliers and the $inlier \%$ of which are approximately the same as the average, rather than use the scans with more inliers and a higher $inlier \%$ for our method. Then we extract a similar number of ISS and Sift3D keypoints for comparison methods. The used scan ID and the specific number of edge keypoints are shown in \cref{tab:Num}. The matching results of our method on each scene are shown in \cref{fig6}, and the comparison results on different scenes are shown in \cref{tab:II} and \cref{tab:III}.

From \cref{tab:II} and \cref{tab:III}, we can see that our method obtains more inliers on each scene than comparison methods. On the challenging forest scene, even though the edge features are unobvious and have a bad effect on the $inliers \%$ of our method, the number of inliers generated by LinK3D is still considerable. On the challenging dynamic scene, the highly dynamic carrier has a bad effect on our method, which reduces the $inlier \%$ for our method. It is noted that our method can effectively match the two scans with a large pose transformation (reverse direction loop), which can be seen in \cref{fig6}(d). Moreover, although the comparison methods achieve comparable results on the point clouds collected from 64-beam LiDAR, due to the sparser points collected from 32-beam and 16-beam LiDAR (the sparsity can be seen from \cref{fig6}(f)(g)(h)) on the M2DGR and Steven VLP16 datasets, the number of inliers for most comparison methods is zero, and they may be unreliable on the point clouds generated by 32-beam and 16-beam LiDAR. \par

\begin{table}[!htbp]
	\centering
	\caption{The efficiency comparison of different methods. The data for \\ 3DFeatNet, 3DSmoothNet, and DH3D is from \cite{du2020dh3d}, and the data \\ for other  DNN-based methods is from their source paper. \\ All units are in seconds.}
	\begin{tabular}{c c|c c c|c}
		\toprule
		\multicolumn{2}{c|}{Method} & $T_{extraction}$ & $T_{matching}$ & $T_{total}$ & \thead{GPU \\ required} \\
		\midrule
		\multirow{7}{*}{\rotatebox{90}{DNN-based}} & 3DFeatNet \cite{yew20183dfeat} & 0.928 & - & 0.928 & \Checkmark\\
		& 3DSmoothNet \cite{gojcic2019perfect} & 0.414 & - & 0.414 & \Checkmark\\
		& DH3D \cite{du2020dh3d} & 0.080 & - & 0.080 & \Checkmark\\
		& FCGF \cite{choy2019fully} & 0.360 & - & 0.360 & \Checkmark\\
		& D3Feat \cite{bai2020d3feat} & 0.130 & - & 0.130 & \Checkmark\\		
		& StickyPillars \cite{fischer2021stickypillars} & \textbf{0.015} & 0.101 & 0.116 & \Checkmark\\
		\midrule
		\multirow{8}{*}{\rotatebox{90}{Hand-crafted}} & PFH \cite{Rusu2008} & 11.148 & 0.077 & 11.225 & \XSolidBrush\\		
		& FPFH \cite{Rusu2009} & 5.601 & 0.015 & 5.616 & \XSolidBrush\\
		& 3DSC \cite{Frome2004} & 0.023 & 7.106 & 7.129 & \XSolidBrush\\
		& SHOT \cite{Tombari2010} & 0.490 & 2.206 & 2.696 & \XSolidBrush\\
		& BSHOT \cite{prakhya2015b} & 0.574 & 0.057 & 0.631 & \XSolidBrush\\
		& 3DHoPD \cite{Prakhya2017a} & 0.414 & \textbf{0.005} & 0.419 & \XSolidBrush\\
		& LinK3D (ours) & 0.030 & 0.020 & \textbf{0.050} & \XSolidBrush\\
		\bottomrule
	\end{tabular}
	\label{tab:realtime comparison}
\end{table}

\subsection{Real-Time Performance Evaluation}
 In this section, we evaluate the efficiency of our method and compare it with the DNN-based and handcrafted methods. KITTI 00 is used for the evaluation of handcrafted methods, which includes 4500+ LiDAR scans, and each LiDAR scan contains approximately 120000+ points. We compute the average runtime for our method when the sequential LiDAR scans are matched. We compute the average runtime after multiple measurements on the used city scene in \cref{matching performance} for other hand-crafted methods. For DNN-based methods, the runtime of StickyPillars is evaluated on KITTI, and others are evaluated on 3DMatch \cite{zeng20173dmatch}, which is collected from an RGB-D camera and each frame of which contains approximately 27000+ points ($<$ KITTI's) on average. \cref{tab:realtime comparison} shows the comparison results. We can see that the DNN methods usually take more time, although fewer points are used and GPUs are required. Other handcrafted methods cannot achieve the real-time performance. LinK3D only takes about 50 milliseconds to extract and match features on average, and it shows great real-time performance.

\begin{table}[!htbp]
	\centering
	\caption{The success rate of LinK3D in unstructured high-way and forest scenes on the KITTI dataset.}
	\begin{tabular}{ c c c c}
		\toprule
		& KITTI 01 & KITTI 02 & KITTI 09 \\
		\midrule
		Success Rate (\%) & 88.6 & 97.6 & 99.9 \\
		\bottomrule
	\end{tabular}
	\label{tab:MatchingFailure}
\end{table}

\subsection{Matching Failure Case on Unstructured Scenes}
When we evaluated our algorithm on KITTI (from sequence 00 to 10), we found that our method may fail to generate true matches between two sequential LiDAR scans after RANSAC in the unstructured KITTI 01, 02, and 09 with fewer valid edge features. KITTI 01 is collected from a high-way scene, and some LiDAR scans of KITTI 02 and 09 are collected from the forest scene. We use the ground truth to determine the success rate on the three sequences, which is shown in \cref{tab:MatchingFailure}. The results indicate that our method may not be robust for the unstructured scenes with fewer effective edge features. In the future, we will continue to improve the robustness of our method for unstructured scenes.

\subsection{Application: LiDAR Odometry}
LiDAR odometry is usually the front-end of the LiDAR SLAM system, and the estimated results of LiDAR odometry have an effect on the accuracy of the SLAM back-end. In this part, we embed LinK3D in combination with a subsequent mapping step. For this purpose, we replace the scan-to-scan registration step of A-LOAM\footnotemark[3]\footnotetext[3]{https://github.com/HKUST-Aerial-Robotics/A-LOAM} with the registration results of SVD solution \cite{arun1987least} based on the matching results of LinK3D. The original registration results of A-LOAM are used when LinK3D encounters matching failure cases. We compare our method with CLS \cite{velas2016collar}, LOAM \cite{zhang2014loam} and LO-Net \cite{li2019net} in \cref{tab:odometry}, and use the KITTI odometry metrics \cite{Geiger2012} to quantitatively analyze the accuracy of LiDAR odometry. The results are shown in \cref{tab:odometry}. It can be seen that the odometry based on LinK3D achieves comparable estimation results.   

\begin{table}[!htbp]
	\centering
	\caption{The comparison results of LiDAR odometry on KITTI dataset.}
	\begin{tabular}{c | c c c c c c | c c}
		\toprule
		\multirow{2}{*}{Sequence} & \multicolumn{2}{c}{CLS \cite{velas2016collar}} & \multicolumn{2}{c}{LOAM \cite{zhang2014loam}} & \multicolumn{2}{c|}{LO-Net \cite{li2019net}}  & \multicolumn{2}{c}{Ours} \\
		&$t_{rel}$ & $r_{rel}$ & $t_{rel}$ & $r_{rel}$ & $t_{rel}$ & $r_{rel}$ & $t_{rel}$ & $r_{rel}$ \\
		\midrule
		00 & 2.11 & 0.95 & 0.78 & 0.53 & 0.78 & 0.42  & \textbf{0.71} & \textbf{0.32}\\
		01 & 4.22 & 1.05 & 1.43 & 0.55 & \textbf{1.42} & \textbf{0.40}  & 4.81 & 0.50\\
		02 & 2.29 & 0.86 & \textbf{0.92} & 0.55 & 1.01 & 0.45 & 1.33 & \textbf{0.40}\\
		03 & 1.63 & 1.09 & 0.86 & 0.65 & \textbf{0.73} & \textbf{0.59} & 1.39 & 0.64\\
		04 & 1.59 & 0.71 & 0.71 & 0.50 & \textbf{0.56} & 0.54 & 0.57 & \textbf{0.40} \\
		05 & 1.98 & 0.92 & \textbf{0.57} & 0.38 & 0.62 & 0.35 & 0.63 & \textbf{0.27} \\
		06 & 0.92 & 0.46 &0.65 & 0.39 & \textbf{0.55} & 0.33 & 0.64 & \textbf{0.28} \\
		07 & 1.04 & 0.73 & 0.63 & 0.50 & \textbf{0.56} & 0.45 & 0.63 & \textbf{0.26} \\
		08 & 2.14 & 1.05 & 1.12 & 0.44 & 1.08 & 0.43 & \textbf{1.03} & \textbf{0.32}\\
		09 & 1.95 & 0.92 & 0.77 & 0.48 & 0.77 & 0.38 & \textbf{0.74} & \textbf{0.29} \\
		10 & 3.46 & 1.28 & \textbf{0.79} & 0.57 & 0.92 & \textbf{0.41} & 1.17 & 0.42 \\
		Mean & 2.12 & 0.91 & 0.84 & 0.51 & \textbf{0.82} & 0.43 & 1.24 & \textbf{0.37}\\
		\bottomrule
	\end{tabular}
	\label{tab:odometry}
\end{table}

\section{Conclusion and Future work} \label{conclusion}
In this work, we propose a novel 3D feature representation method to solve the issue that the existing methods are not fully applicable to the sparse 3D LiDAR point cloud, which is called LinK3D. The core idea of LinK3D is to describe the current keypoint with its neighboring keypoints. The experimental results show that LinK3D can generate a large number of valid matches on sparse 16-, 32-, and 64-beam LiDAR point clouds in real time. In the future, we will continue to improve the robustness of LinK3D for unstructured scenes. It is promising to extend LinK3D into more possible 3D vision tasks (e.g., place recognition and relocalization of mobile robots) to improve the efficiency and accuracy of different LiDAR-based mobile robot systems.

\bibliographystyle{IEEEtran}
\bibliography{MyRef}

\end{document}